\pdfoutput=1

\documentclass[11pt]{article}
\usepackage{colortbl}
\usepackage{amsmath} 
\usepackage{todonotes}
\usepackage{tablefootnote}
\usepackage{float}
\usepackage{textcomp}
\usepackage{multirow}
\usepackage[final]{acl}
\usepackage{times}
\usepackage{latexsym}
\usepackage{adjustbox} 

\usepackage[font=small]{caption} 

\usepackage[T1]{fontenc}

\usepackage[utf8]{inputenc}
\usepackage{hyperref}

\usepackage{microtype}
\usepackage{booktabs}
\usepackage{inconsolata}

\usepackage{graphicx}
\usepackage{tabularx} 

\newcommand{\ignore}[1]{}
\title{When Does Language Transfer Help? Sequential Fine-Tuning for Cross-Lingual Euphemism Detection}

\author{Julia Sammartino, {\bf Libby Barak }, {\bf Jing Peng}, {\bf Anna Feldman} \\
      Montclair State University\\New Jersey, USA\\
  \small\texttt{\{sammartinoj1, barakl, pengj, feldmana\}@montclair.edu}}

\begin{document}
\maketitle
\begin{abstract}


Euphemisms are culturally variable and often ambiguous, posing challenges for language models, especially in low-resource settings. This paper investigates how cross-lingual transfer via sequential fine-tuning affects euphemism detection across five languages: English, Spanish, Chinese, Turkish, and Yorùbá. We compare sequential fine-tuning with monolingual and simultaneous fine-tuning using XLM-R and mBERT, analyzing how performance is shaped by language pairings, typological features, and pretraining coverage. Results show that sequential fine-tuning with a high-resource L1 improves L2 performance, especially for low-resource languages like Yorùbá and Turkish. XLM-R achieves larger gains but is more sensitive to pretraining gaps and catastrophic forgetting, while mBERT yields more stable, though lower, results. These findings highlight sequential fine-tuning as a simple yet effective strategy for improving euphemism  detection in multilingual models, particularly when low-resource languages are involved.

\end{abstract}

\section{Introduction}

Euphemisms are used as substitutes for words or phrases that could be considered harsh, impolite, or taboo. For example, instead of overtly stating that someone died, one can instead utilize a euphemism that softens the tone: "I heard that his dad \textbf{passed on}." Due to the subjectiveness and figurative nature of euphemisms, native speakers of a language may disagree on whether a word or phrase is considered euphemistic \cite{gavidia-2022-cats}. It is important to note that some phrases which may be used as euphemisms may also be taken at face value in certain contexts, without an underlying intended meaning (e.g. "he \textbf{passed on} the information to his boss", "I \textbf{passed on} this job offer"). Therefore, the term Potentially Euphemistic Terms (PETs) was created to reflect this ambiguity, aligning with previous research \cite{gavidia-2022-cats,lee-etal-2022-searching}. For example, the phrase 'between jobs' could be used euphemistically to mean 'unemployed', or could be taken literally to mean 'between shifts at two different jobs'. We investigate whether models can transfer euphemism knowledge across languages, and whether sequential fine-tuning—training first on a high-resource language—can improve performance in low-resource settings. 

Multilingual transformer models, such as XLM-RoBERTa (XLM-R) \cite{xlmr} and mBERT \cite{mbert} have been used for various tasks and experiments due to their ability to capture cross-lingual representations and transfer-learning capabilities. In order to analyze the knowledge captured through this process, we performed experiments with sequential fine-tuning and compared it to monolingual baselines and paired language simultaneous fine-tuning. We investigate the cross-linguistic generalization capabilities through sequential fine-tuning, in which the model learns the same task first on one language, L1, and once it reaches its peak performance, learns the same task on a second language, L2. The model is then tested on both languages, including every pairing of English (EN), Mandarin Chinese (ZH), (Latin American and Castilian) Spanish (ES), Turkish (TR), and Yorùbá (YO). 


We then compare these sequential results to the baseline monolingual score for each language, as well as simultaneous fine-tuning - learning two languages at the same time, and then testing on both. 


We hypothesize that these proposed experimental settings of sequential fine-tuning will enable deeper understanding of the abilities of LLMs to learn the properties of abstract figurative language when given the chance to focus on each language in isolation. This experiment is especially important for low-resource languages, in which we have less access to rich training data and therefore depend on cross-lingual transfer to boost a model's performance. Our extensive analysis offers a new perspective on this important aspect of multilingual LLMs, and investigates the cross-lingual capabilities of XLM-R and mBERT.

\begin{table}[htbp]
\begin{center}
\scalebox{0.8}{
\begin{tabular}{|ccc|c|} 
    \hline
    \textbf{Lang} & \textbf{Euph} & \textbf{Non-Euph} & \textbf{Total}  \\
    \hline
    ZH  & 2213 (149) & 998 (56)  & 3211 (151) \\
    EN & 1841 (141) & 1257 (85) & 3098 (144) \\
    ES & 1955 (223) & 997 (135) & 2952 (233) \\
    TR & 1457 (67) & 979 (59) & 2436 (70)  \\
    YO & 1689 (153) & 909 (85) & 2598 (157) \\
    \hline
  \end{tabular}
  }
  \caption{Number of examples for 2025 PETs Datasets - Number of PETs for each class in parentheses. For each individual PET, there is a maximum of 40 examples of each class (euph vs. non-euph).}
  \label{tab:pets-overall2}
  \end{center}
\end{table}

\section{Related Work}

\subsection{Euphemism Detection}

Recent work on euphemism detection has expanded to multilingual settings, leveraging deep learning and cross-lingual methods. \citet{gavidia-2022-cats} introduced a PETs corpus for English, later extended to Spanish, Chinese, Yorùbá \cite{lee-etal-2023-feed,lee-etal-2024-meds} and Turkish \cite{biyik-etal-2024-turkish}.

Approaches range from lexicon-based methods \cite{felt-riloff-2020-recognizing,lee-etal-2022-searching} to transformer models \cite{zhu-2021-self-supervised,wang2022euphemism} to exploring various linguistic properties, e.g., vagueness \cite{lee-etal-2023-feed}. To address data scarcity, \citet{kohli2022adversarial} used adversarial augmentation, and \citet{keh2022eureka} applied kNN-based data expansion. Shared tasks \cite{lee2022report,lee2024multilingual} have driven benchmarking, with ensemble models \cite{vitiugin2024ensemble} achieving strong performance, while zero-shot evaluations \citet{keh2022exploring} provide insights into cross-lingual generalization.

Multilingual work with more than two languages remains limited, though bilingual euphemism detection \cite{wang2022euphemism} and euphemistic abuse detection \cite{wiegand2023euphemistic} highlight transfer challenges. We extend this by evaluating sequential fine-tuning across diverse languages, analyzing the role of dataset structure and lexical overlap in cross-lingual transfer.

\subsection{Sequential Fine-Tuning}

Sequential fine-tuning is an established approach to LLM experimentation, but the majority of previous work has focused on utilizing multiple tasks or `sub-tasks' rather than exploring the cross-lingual capabilities of a model. 
Prior work has shown that continued pretraining on domain- or task-specific data improves downstream performance, even without labeled supervision \citet{gururangan-etal-2020-dont}. Our study builds on this insight, applying a similar principle to the cross-lingual setting, where we use labeled data for figurative language (euphemisms) in one language as a form of task-aligned adaptation for another.
One cross-lingual application focused on translation into English and then classification \cite{hu2024finetuninglargelanguagemodels}. This highlights a downside to some multilingual work with LLMs -- having the model work on a dataset that has been translated into English, rather than directly interpreting the original non-English text. Our work improves upon this area by using a variety of languages \textbf{without} using the model for translation, as translation may result in a loss of underlying meanings. 

Figurative language adds another layer of complexity as far as underlying meanings. Prior work has evaluated euphemism detection in multilingual or zero-shot settings \citet{lee-etal-2023-feed, keh2022exploring}, but few studies have tested sequential fine-tuning as a method for targeted cross-lingual adaptation. We address this gap by systematically evaluating whether exposure to euphemisms in a high-resource language improves detection in a low-resource language, using both mBERT and XLM-R across five typologically diverse languages.

\section{Datasets}
We leverage publicly available euphemism datasets that were originally published in \citet{lee2024multilingual}, with the addition of a Turkish dataset. Table \ref{tab:pets-overall2} details the distribution of euphemistic and non-euphemistic examples.

Previous researchers created these datasets by first curating a list of potentially euphemistic terms, and then scraping from a variety of corpora that are listed in the following paper \cite{lee-etal-2023-feed}. This data is composed of extracted examples from online sources including Glowbe (English) \cite{davies2013glowbe} and curated corpora (Spanish, Chinese)  \cite{rae2025corpes, brightmart2019nlp}.  In the case of Yorùbá and Turkish \footnote{The Turkish dataset was used with permission from the author for a paper that is currently under review. The curation of it followed a similar schematic to previous work.}, the authors utilized various sources such as news articles, religious texts, and more.  

Annotations were executed by at least 3 native speakers of each language, and majority vote was utilized for the final classification. \citet{lee-etal-2024-meds} assessed inter-annotator agreement using Krippendorf's alpha on a small subset of the dataset, and found values ranging from 0.415 to 0.679 on a scale of 0 to 1. This was expected, as euphemisms can be ambiguous, even to native speakers.

\section{Methodology}
\subsection{Model}
As our research focused on the multilingual and cross-lingual learning capabilities of LLMs, we chose to experiment with two prominent multilingual models - XLM-R and mBERT.

XLM-R was pretrained on English, Chinese, Spanish, and Turkish, but not on Yorùbá. The model was trained on 2.5TB of CommonCrawl data spanning 100 languages, notably with English, Chinese, and Spanish receiving significantly higher representation, Turkish having moderate coverage, and Yorùbá absent from the pretraining corpus. XLM-R has approximately 125 million trainable parameters with 12 hidden layers with 768-dimensional hidden states \cite{xlmr}.

mBERT was pretrained on Wikipedia data for 104 languages - with the explicit caveat that lower resource languages have less training data overall. For our experimentation, however, all five languages are included in pretraining, a major difference from XLM-R. This model has 110 million trainable parameters and 12 hidden layers with 768-dimensional hidden states, making its size and fine-tuning capabilities comparable to XLM-R \cite{mbert}. \footnotetext{Pretraining coverage for mBERT can be found \href{https://github.com/google-research/bert/blob/master/multilingual.md}{here} and for XLM-R \href{https://arxiv.org/pdf/1911.02116}{here}.}


\begin{table}[ht]
\begin{center}
\setlength{\arrayrulewidth}{1pt}
\scalebox{0.8}{
\begin{tabular}{|c|c|c|c|c|c|} 
 \hline
 \textbf{Model} & \textbf{EN} & \textbf{ES} & \textbf{ZH} & \textbf{YO} & \textbf{TR}  \\
 \hline
  \text{XLM-R} & 0.821  & 0.768 & 0.878 & 0.809 & 0.790  \\
  \text{mBERT} & 0.791 & 0.712 & 0.860 & 0.800 & 0.720 \\
 \hline
\end{tabular}
}
\caption{Average Macro-F1s for Monolingual Fine-Tuning}
\label{tbl:mono}
\end{center}
\end{table}

\subsection{Experimental Setup}
Each experiment consisted of five trials with an 80-10-10 training-validation-testing split. Datasets included a `euph\_status' feature distinguishing always-euphemistic from sometimes-euphemistic PETs. To prevent memorization, always-euphemistic PETs appeared only in training or testing, ensuring the model learned euphemism use from context. We used standard hyperparameters, including a $1 \times 10^{-5}$ learning rate, AdamW optimizer, and batch size of 4. Fixed learning rate experiments performed similarly or worse. Models were trained for up to 15 epochs with early stopping (patience = 5). Training on GPUs took $\sim$6 hours for sequential fine-tuning and $\sim$5 hours for simultaneous fine-tuning per language pair.

These experiments are designed to test the ability to transfer knowledge of euphemisms learned in one language to another language. To assess the directionality of transfer, all language pairs are evaluated bidirectionally (e.g., English $\rightarrow$ Yorùbá and Yorùbá $\rightarrow$English), allowing us to analyze both symmetric and asymmetric patterns of cross-lingual adaptation.

\renewcommand{\thefootnote}{}
\footnotetext{\href{https://github.com/sammartinoj/PETScan}{Github for Code and Data}}
\begin{table*}[ht]
\begin{center}
\setlength{\arrayrulewidth}{1pt}
\begin{tabular}{|l|c|c|c|c|} 
 \hline
 \multirow{2}{*}{\textbf{Lang. Pair}} & \multicolumn{2}{c|}{\textbf{XLM-R}} & \multicolumn{2}{c|}{\textbf{mBERT}} \\
 \cline{2-5}
  & \textbf{Lang A } & \textbf{Lang B} & \textbf{Lang A} & \textbf{Lang B} \\
 \hline
 \textbf{EN \& ES} & 0.821 (0.821) & \underline{0.781} (0.768) & 0.801 (0.791) & \underline{0.733} (0.712) \\
 \textbf{EN \& ZH} & \underline{0.829} (0.821) & \underline{0.885} (0.878) & \underline{0.808} (0.791) & 0.852 (0.860)\\ 
 \textbf{EN \& YO} &\underline{0.829} (0.821) & 0.455 (0.809) & 0.789 (0.791) & \underline{0.814} (0.800)\\ 
 \textbf{EN \& TR} & \underline{0.832} (0.821) & \underline{0.817} (0.790) & \underline{0.803} (0.791) & \underline{0.759} (0.720) \\ 
 \textbf{ES \& ZH} & 0.768 (0.768) & \underline{0.893} (0.878) & \underline{0.732} (0.712) & 0.850 (0.860) \\
 \textbf{ES \& YO} & 0.741 (0.768) & 0.797 (0.809) & \underline{0.728} (0.712) & 0.800 (0.800)\\ 
 \textbf{ES \& TR} & 0.751 (0.768) & \underline{0.802} (0.790) & 0.700 (0.712) & \underline{0.731} (0.720) \\ 
 \textbf{ZH \& YO} & \underline{0.882} (0.878) & \underline{0.824} (0.809) & 0.855 (0.860) & \underline{0.808} (0.800)\\
 \textbf{ZH \& TR} & 0.873 (0.878) & \underline{0.808} (0.790)  & 0.831 (0.860) & \underline{0.747} (0.720)\\ 
 \textbf{YO \& TR} & \underline{0.811} (0.809) & \underline{0.795} (0.790) & 0.793 (0.800) & \underline{0.729 (0.720)}\\ 
 \hline
\end{tabular}
\caption{Average Macro-F1s for Simultaneous Fine-Tuning. Monolingual (Baseline) scores are reported in parentheses. F1 scores outperforming the baseline are underscored.}
\label{tab:simul-results}
\end{center}
\end{table*}

\section{Simultaneous and Sequential Fine-Tuning Results} \label{results}
\subsection{Baseline}
We maintain a consistent parameter setting for the monolingual experiments as done for sequential and simultaneous fine-tuning models. We observe a decrease in training time due to the relatively smaller size of the training data. As shown in Table \ref{tbl:mono}, XLM-R consistently outperforms mBERT, possibly due to the slight difference in number of trainable parameters.

\vspace{-3mm}
\begin{table*}[t]
\begin{center}
\small 
\setlength{\arrayrulewidth}{1pt}
\begin{tabular}{|c|c|c|c|c|c|} 
 \hline
  \multicolumn{2}{|c|}{\textbf{Train}} & \multicolumn{2}{c|}{\textbf{Test - XLM-R}} & \multicolumn{2}{c|}{\textbf{Test - mBERT}}\\
  \hline
  \textbf{L1} & \textbf{L2} & \textbf{L1} & \textbf{L2} & 
  \textbf{L1} & \textbf{L2}\\
 
 \hline
 \text{ES} & \text{EN} & 0.733 (0.768) & \cellcolor{cyan!15} \underline{0.824} (0.821) & 0.702 (0.712) & \cellcolor{cyan!15}\underline{0.799} (0.791) \\
 \text{ZH} & \text{EN} &  0.791 (0.878) &\cellcolor{cyan!15} \underline{0.830} (0.821) & 0.809 (0.860) & \cellcolor{cyan!15}\underline{0.812} (0.791) \\ 
 \text{YO} & \text{EN} &  0.490 (0.809) &\cellcolor{cyan!15} 0.800 (0.821) & 0.678 (0.800) & \cellcolor{cyan!15}0.785 (0.791)\\ 
 \text{TR} & \text{EN} &  0.732 (0.790) &\cellcolor{cyan!15} \underline{0.835} (0.821) & 0.660 (0.720) & \cellcolor{cyan!15}0.791 (0.791)\\ 
 
 \hline
 \text{EN} & \text{ES} & \cellcolor{yellow!35} 0.780 (0.821) & 0.761 (0.768) & \cellcolor{yellow!35}0.745 (0.791) & \underline{0.738} (0.712) \\
 \text{ZH} & \text{ES} &\cellcolor{yellow!35} 0.843 (0.878) & 0.746 (0.768) & \cellcolor{yellow!35}0.843 (0.860) & \underline{0.722} (0.712)\\ 
 \text{YO} & \text{ES} &  0.709 (0.809) & \cellcolor{cyan!15}0.746 (0.768) & \cellcolor{yellow!35} 0.770 (0.800) & \underline{0.717} (0.712) \\ 
 \text{TR} & \text{ES} &  0.676 (0.790) &\cellcolor{cyan!15} 0.764 (0.768) & 0.622 (0.720) & \cellcolor{cyan!15}0.690 (0.712)\\ 
 
 \hline
 \text{EN} & \text{ZH} & 0.797 (0.821) & \cellcolor{cyan!15} 0.876 (0.878) & 0.783 (0.791) & \cellcolor{cyan!15}\underline{0.868} (0.860)\\
 \text{ES} & \text{ZH} & 0.743 (0.768) &\cellcolor{cyan!15} 0.876 (0.878) &\underline {0.727} (0.712) & \cellcolor{cyan!15}\underline{0.885} (0.860)\\ 
 \text{YO} & \text{ZH} &  0.432 (0.809) &\cellcolor{cyan!15} \underline{0.900} (0.878) & 0.701 (0.800) &\cellcolor{cyan!15} 0.854 (0.860)\\ 
 \text{TR} & \text{ZH} &  0.704 (0.790) & \cellcolor{cyan!15}0.857 (0.878) & 0.676 (0.720) & \cellcolor{cyan!15}0.858 (0.860)\\ 
 
 \hline
 \text{EN} & \text{YO} & 0.761 (0.821) & \cellcolor{cyan!15} \underline{0.812} (0.809) & 0.735 (0.791) & \cellcolor{cyan!15}\underline{0.817} (0.800)\\
 \text{ES} & \text{YO} & 0.661 (0.768) &\cellcolor{cyan!15} \underline{0.830} (0.809) & \underline{0.734} (0.712) & \cellcolor{cyan!15}\underline{0.801} (0.800) \\ 
 \text{ZH} & \text{YO} &\cellcolor{yellow!35} 0.837 (0.878) & 0.798 (0.809) & \cellcolor{yellow!35}0.827 (0.860) & \underline{0.809} (0.800) \\ 
 \text{TR} & \text{YO} &  0.727 (0.790) &\cellcolor{cyan!15} \underline{0.824} (0.809) & 0.703 (0.720) & \cellcolor{cyan!15}\underline{0.816} (0.800) \\ 
 
 \hline
 \text{EN} & \text{TR} & 0.767 (0.821) & \cellcolor{cyan!15} \underline{0.801} (0.790) & 0.765 (0.791) & \cellcolor{cyan!15}\underline{0.780} (0.720)\\
 \text{ES} & \text{TR} & 0.644 (0.768) & \cellcolor{cyan!15}0.777 (0.790) & 0.662 (0.712) & \cellcolor{cyan!15}\underline{0.741} (0.720)\\ 
 \text{ZH} & \text{TR} &  0.692 (0.878) &\cellcolor{cyan!15} \underline{0.792} (0.790) & 0.727 (0.860) &\cellcolor{cyan!15} \underline{0.758} (0.720)\\ 
 \text{YO} & \text{TR} &  0.674 (0.809) & \cellcolor{cyan!15}0.776 (0.790) &\cellcolor{yellow!35}0.760 (0.800) & \underline{0.742} (0.720) \\ 
 \hline
\end{tabular}
\caption{Comparison of Average Macro-F1s for Sequential Fine-Tuning vs. Monolingual Baseline. Parentheses contain monolingual F1 for reference. F1-scores that outperform the baseline are underscored. Scores highlighted in blue are where L2 performs better than L1, those highlighted in yellow are where L1 outperforms.}
\label{tbl:combined}
\end{center}
\end{table*}

\subsection{Simultaneous Fine-Tuning}

Simultaneous fine-tuning involves combining two languages' datasets for training and validation, while testing on each language separately. Along with the addition of Turkish \cite{biyik-etal-2024-turkish}, our experimentation differs from prior work by using the aforementioned euphemism status-based zero-shot setting, where datasets are shuffled without designating an L1 or L2. Table \ref{tbl:combined} reports the results. 

For XLM-R, Chinese performed well across most pairs (see Table \ref{tab:simul-results}), likely due to strong pretraining data and corpus quality. The only minor drop occurred when paired with Turkish, but it was negligible. Turkish showed stable performance, suggesting compatibility with other languages. Yorùbá struggled in some cases, especially with English, likely due to limited pretraining and English's dominance in XLM-R. Spanish benefited from typologically similar pairs, while English showed mixed results depending on its counterpart. These findings highlight that typological similarity, dataset composition, and pretraining exposure all impact multilingual, simultaneous fine-tuning effectiveness.

mBERT's results for simultaneous fine-tuning were not as prominently different from a language's corresponding baseline as seen in the results for XLM-R.


\begin{figure}[h!]
     \centering
     \includegraphics[width=0.5\textwidth]{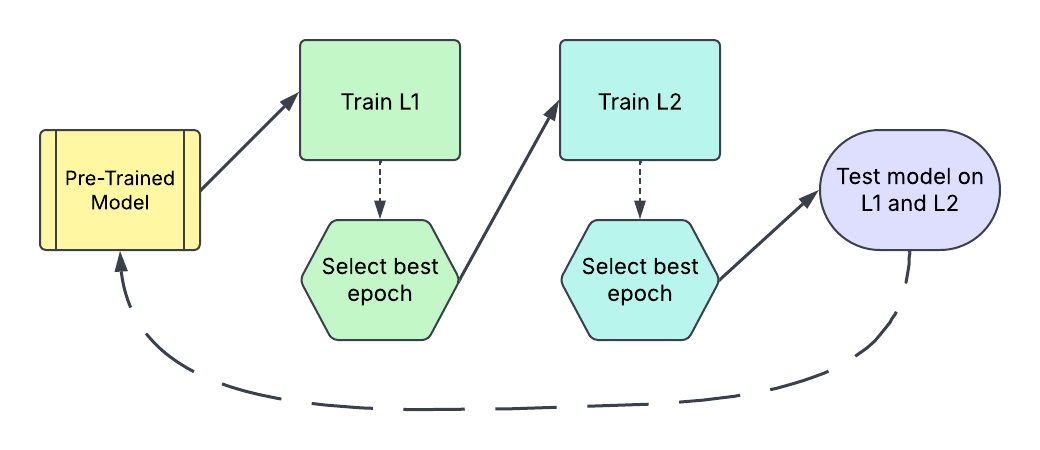}
     \caption{Model Archetype}
     \label{fig:flowchart}
 \end{figure}

\subsection{Sequential Fine-Tuning}

Sequential fine-tuning starts with an off-the-shelf model, which is first fine-tuned on a source language (L1) and then on a target language (L2). The best epoch (based on validation macro F1) on L1 is then fine-tuned on L2. Once the model reaches its highest validation F1 for L2, it is tested on both languages. This process is repeated with a new model for each trial split, and final F1 scores are averaged. The full setup is illustrated in Figure~\ref{fig:flowchart}, and is performed for both XLM-R and mBERT.



\subsubsection{When Does Sequential Fine-Tuning Help L2 Performance?}

Sequential fine-tuning for the majority of experiments with both models reported higher scores for the L2, supporting our original hypothesis that prior knowledge of euphemisms in L1 aids understanding in L2. This setup shows more distinct differences than simultaneous fine-tuning, where scores remained closer to the monolingual baseline.

Examples from Table~\ref{tbl:combined} (XLM-R):
\begin{itemize}\itemsep-1pt
    \item \textbf{EN~$\rightarrow$~YO}: 0.812 vs. YO baseline 0.809
    \item \textbf{ES~$\rightarrow$~YO}: 0.830 vs. 0.809
    \item \textbf{EN~$\rightarrow$~TR}: 0.801 vs. TR baseline 0.790
    \item \textbf{YO~$\rightarrow$~ZH}: 0.900 vs. ZH baseline 0.878
\end{itemize}

These gains are particularly notable in low-resource target languages like Yorùbá and Turkish, reinforcing the benefit of high-resource L1 transfer.

\subsubsection{When Does Sequential Fine-Tuning Hurt L1 Performance? (Catastrophic Forgetting)}

In most cases, L1 performance drops after L2 training—particularly in XLM-R, with some cases resulting in severe performance degradation indicating catastrophic forgetting. This is evident in:

\begin{itemize}\itemsep-1pt
    \item \textbf{YO~$\rightarrow$~EN}: 0.490 vs. EN baseline 0.821
    \item \textbf{YO~$\rightarrow$~ZH}: 0.432 vs. ZH baseline 0.878
\end{itemize}

These considerable drops are not observed in mBERT, likely due to balanced pretraining coverage for all five languages. XLM-R lacks pretraining exposure to Yorùbá, which leads to shallow integration that is easily overwritten. Agreement metrics support this interpretation: Cohen's Kappa between YO monolingual and YO~$\rightarrow$~ZH is 0.145, while YO monolingual and ZH~$\rightarrow$~YO yields 0.667.

\subsubsection{What Roles Do Pretraining, Typology, and Data Play?}

Pretraining coverage significantly impacts transfer success. XLM-R achieves stronger L2 gains but suffers more from volatility, while mBERT shows steadier though lower performance. Yorùbá performs well as L2 but poorly as L1, which tracks with its absence from XLM-R's pretraining corpus.

Typological similarity alone does not explain results. For instance, EN~$\rightarrow$~TR and ES~$\rightarrow$~YO show strong gains despite language distance, whereas EN~$\rightarrow$~ES does not yield consistent improvement.

These findings suggest that dataset characteristics and pretraining exposure are more influential than typological features in euphemism detection transfer.

\subsubsection{Summary of Key Results}
Table \ref{tab:key-results} highlights the top-performing configurations for each language in both models. For XLM-R, the strongest gain occurs in the YO~$\rightarrow$~ZH setting (0.900), outperforming the ZH monolingual baseline (0.878). 

In contrast, mBERT produces more balanced results across languages, with no extreme gains but consistent improvements when English is used as the source language. While mBERT avoids catastrophic forgetting due to its more uniform pretraining coverage, XLM-R achieves higher absolute L2 performance, especially for low-resource L2s. These results underscore that sequential fine-tuning is a lightweight and effective strategy for improving euphemism detection across typologically diverse and unevenly resourced languages.


\vspace{-3mm}
\begin{table}[htbp]
\small
\centering
\begin{tabular}{|l|c|c|c|c|}
\hline
\textbf{Model} & \textbf{Lang.} & \textbf{Pair} & \textbf{Type} & \textbf{F1} \\
\hline
 XLM-R & EN & TR~$\rightarrow$~EN & Seq. & 0.835 \\
 XLM-R & ES & ES \& ZH & Sim. & \text{0.768*} \\
 XLM-R & ZH & YO~$\rightarrow$~ZH & Seq. & 0.9 \\
 XLM-R & YO & ES~$\rightarrow$~YO & Seq. & 0.830 \\
 XLM-R & TR & EN \& TR & Sim. & 0.817 \\
 \hline
 mBERT & EN & ZH~$\rightarrow$~EN & Seq. & 0.812 \\
 mBERT & ES & EN~$\rightarrow$~ES & Seq. & 0.738 \\
 mBERT & ZH & ES~$\rightarrow$~ZH & Seq. & 0.885 \\
 mBERT & YO & EN~$\rightarrow$~YO & Seq. & 0.817 \\
 mBERT & TR & EN~$\rightarrow$~TR & Seq. & 0.780\\
\hline
\end{tabular}
\caption{Highest F1 scores for models in each of the languages over the two fine-tuning setups: sequential (Seq.) and simultaneous (Sim.). * indicates score matches baseline performance.}
\label{tab:key-results}
\end{table}


\section{Conclusion and Future Work}
This paper explored whether euphemism detection can benefit from cross-lingual transfer, specifically through sequential fine-tuning. We evaluated XLM-R and mBERT across five typologically and resource-diverse languages: English, Spanish, Chinese, Turkish, and Yorùbá.

Our findings show that sequential fine-tuning with a high-resource language improves L2 euphemism detection, especially for low-resource languages like Yorùbá and Turkish. XLM-R achieves larger gains, but is more sensitive to catastrophic forgetting and pretraining gaps. mBERT, by contrast, shows more stable performance across language pairs, albeit with smaller improvements.

Interestingly, the success of transfer was not predicted by typological similarity. Instead, performance was shaped more by dataset structure and pretraining exposure. Strong results for Yorùbá$\rightarrow$Chinese and English$\rightarrow$Turkish demonstrate that meaningful transfer can occur even between distant languages.

Overall, these results highlight sequential fine-tuning as a lightweight and effective adaptation strategy for figurative language tasks and extends previous studies by introducing cross-lingual transfer investigations in relation to a challenging task. In future work, we plan to explore few-shot sequential fine-tuning, hybrid multilingual-sequential setups, and extensions to languages with non-Latin scripts and richer morphology.


Future work could explore cyclical fine-tuning or interleaved exposure to counteract forgetting, and longer L2 training where L1 outperforms. Testing whether Yorùbá’s weaker performance extends to other low-resource languages could reveal if sequential fine-tuning serves as implicit pretraining.

Evaluating larger multilingual models (e.g., mT5, GPT-4, Mistral) may enhance cross-lingual euphemism detection, particularly for low-resource languages. Expanding to morphologically rich and non-Latin scripts could uncover new challenges, while discourse-level modeling may improve context sensitivity. This study shows that prior exposure to euphemisms in L1 enhances cross-lingual transfer, but effectiveness depends on pretraining data, dataset structure, and linguistic differences. Sequential fine-tuning provides a scalable strategy for improving LLM's ability to detect figurative language in low-resource settings, thus contributing to the development of more effective multilingual NLP models.

\section*{Limitations}

Our study has several limitations in cross-lingual euphemism detection. Dataset imbalance affects comparability, as Spanish contains significantly more PETs than other languages, which may skew model performance. XLM-R’s pretraining bias favors English, Spanish, and Chinese, while Turkish has moderate coverage, and Yorùbá has none, contributing to its weaker performance in some settings. Furthermore, most of the datasets were skewed towards the 1's (i.e. euphemistic contexts), with the Chinese dataset and the Spanish dataset having nearly 2/3 of their instances labeled as Euphemistic.

Catastrophic forgetting occurred in sequential fine-tuning with XLM-R, where L1 performance dropped after exposure to L2, particularly in YO~$\rightarrow$~EN and YO~$\rightarrow$~ZH, indicating interference in euphemism learning. Typology did not strictly predict transfer success -- some distant pairs (e.g., EN~$\rightarrow$~TR, ES~$\rightarrow$~YO) showed gains, while structurally similar languages (e.g., English~$\rightarrow$~Spanish) did not, suggesting dataset complexity and euphemism structures play a larger role. 

Computational constraints may have impacted results, but we did not systematically test training duration and batch sizes with larger models. Due to efficiency constraints, we were only able to perform 5 trials on each pair for sequential fine-tuning - although the model still sees the entirety of our datasets, it does not receive as much variability in regards to the random shuffles. 

Generalizability remains uncertain, as all studied languages use relatively simple scripts, with the exception of Chinese, which uses a logographic script, leaving open questions about languages with complex morphology or non-Latin scripts. Finally, euphemism detection is inherently subjective, meaning dataset inconsistencies and cultural variation may introduce noise.

\section*{Ethics Statement}
This study acknowledges the cultural and linguistic variations in euphemism detection, as meanings shift across contexts. The data used in this work was made publicly available by the authors of \citet{lee2024multilingual} and is used in accordance with their original intent. 

Our dataset includes euphemisms related to sensitive topics like death, illness, and socio-political issues, and may include vulgar language. 
We are not policing language -- our goal is to enhance cross-lingual understanding of euphemistic language. The data used does not contain personally identifying information. 

Our research includes low-resource languages like Yorùbá, which often lack strong NLP infrastructure. By working with these languages, we aim to support more inclusive language technologies.

\section*{Acknowledgments}
Thanks to Patrick Lee, Hasan Biyik, and Whitney Poh for help with annotation, experiments, and feedback.
This material is based upon work supported by the National Science Foundation under Grants \#2226006 and \#2428506.


\bibliography{custom}

\begin{thebibliography}{22}
\providecommand{\natexlab}[1]{#1}

\bibitem[{Biyik et~al.(2024)Biyik, Lee, and Feldman}]{biyik-etal-2024-turkish}
Hasan Biyik, Patrick Lee, and Anna Feldman. 2024.
\newblock \href {https://aclanthology.org/2024.sigturk-1.7/} {{T}urkish
  delights: a dataset on {T}urkish euphemisms}.
\newblock In \emph{Proceedings of the First Workshop on Natural Language
  Processing for Turkic Languages (SIGTURK 2024)}, pages 71--80, Bangkok,
  Thailand and Online. Association for Computational Linguistics.

\bibitem[{Brightmart(2019)}]{brightmart2019nlp}
Brightmart. 2019.
\newblock \href {https://doi.org/10.5281/zenodo.3402022} {{NLP Chinese Corpus}:
  Release version 1.0}.
\newblock Accessed via Zenodo.

\bibitem[{Conneau et~al.(2019)Conneau, Khandelwal, Goyal, Chaudhary, Wenzek,
  Guzm{\'{a}}n, Grave, Ott, Zettlemoyer, and Stoyanov}]{xlmr}
Alexis Conneau, Kartikay Khandelwal, Naman Goyal, Vishrav Chaudhary, Guillaume
  Wenzek, Francisco Guzm{\'{a}}n, Edouard Grave, Myle Ott, Luke Zettlemoyer,
  and Veselin Stoyanov. 2019.
\newblock \href {https://arxiv.org/abs/1911.02116} {Unsupervised cross-lingual
  representation learning at scale}.
\newblock \emph{CoRR}, abs/1911.02116.

\bibitem[{Davies(2013)}]{davies2013glowbe}
Mark Davies. 2013.
\newblock Corpus of global web-based english: 1.9 billion words from speakers
  in 20 countries (glowbe).
\newblock \url{https://corpus.byu.edu/glowbe/}.
\newblock Accessed: 2025-05-25.

\bibitem[{Felt and Riloff(2020)}]{felt-riloff-2020-recognizing}
Christian Felt and Ellen Riloff. 2020.
\newblock \href {https://doi.org/10.18653/v1/2020.figlang-1.20} {Recognizing
  euphemisms and dysphemisms using sentiment analysis}.
\newblock In \emph{Proceedings of the Second Workshop on Figurative Language
  Processing}, pages 136--145, Online. Association for Computational
  Linguistics.

\bibitem[{Gavidia et~al.(2022)Gavidia, Lee, Feldman, and
  Peng}]{gavidia-2022-cats}
Martha Gavidia, Patrick Lee, Anna Feldman, and Jing Peng. 2022.
\newblock \href {https://aclanthology.org/2022.lrec-1.285/} {{CAT}s are fuzzy
  {PET}s: A corpus and analysis of potentially euphemistic terms}.
\newblock In \emph{Proceedings of the Thirteenth Language Resources and
  Evaluation Conference}, pages 2658--2671, Marseille, France. European
  Language Resources Association.

\bibitem[{Gururangan et~al.(2020)Gururangan, Marasovi{\'c}, Swayamdipta, Lo,
  Beltagy, Downey, and Smith}]{gururangan-etal-2020-dont}
Suchin Gururangan, Ana Marasovi{\'c}, Swabha Swayamdipta, Kyle Lo, Iz~Beltagy,
  Doug Downey, and Noah~A. Smith. 2020.
\newblock \href {https://doi.org/10.18653/v1/2020.acl-main.740} {Don`t stop
  pretraining: Adapt language models to domains and tasks}.
\newblock In \emph{Proceedings of the 58th Annual Meeting of the Association
  for Computational Linguistics}, pages 8342--8360, Online. Association for
  Computational Linguistics.

\bibitem[{Hu et~al.(2024)Hu, Yu, Chen, and
  Ponti}]{hu2024finetuninglargelanguagemodels}
Hanxu Hu, Simon Yu, Pinzhen Chen, and Edoardo~M. Ponti. 2024.
\newblock \href {https://arxiv.org/abs/2403.07794} {Fine-tuning large language
  models with sequential instructions}.
\newblock \emph{Preprint}, arXiv:2403.07794.

\bibitem[{Keh(2022)}]{keh2022exploring}
Sedrick~Scott Keh. 2022.
\newblock \href {https://doi.org/10.18653/v1/2022.flp-1.24} {{Exploring
  Euphemism Detection in Few-Shot and Zero-Shot Settings}}.
\newblock In \emph{Proceedings of the 3rd Workshop on Figurative Language
  Processing (FLP)}, pages 167--172, Abu Dhabi, United Arab Emirates (Hybrid).
  Association for Computational Linguistics.

\bibitem[{Keh et~al.(2022)Keh, Bharadwaj, Liu, Tedeschi, Gangal, and
  Navigli}]{keh2022eureka}
Sedrick~Scott Keh, Rohit Bharadwaj, Emmy Liu, Simone Tedeschi, Varun Gangal,
  and Roberto Navigli. 2022.
\newblock \href {https://doi.org/10.18653/v1/2022.flp-1.15} {{{EUREKA}:
  {EU}phemism Recognition Enhanced through Knn-based methods and
  Augmentation}}.
\newblock In \emph{Proceedings of the 3rd Workshop on Figurative Language
  Processing (FLP)}, pages 111--117, Abu Dhabi, United Arab Emirates (Hybrid).
  Association for Computational Linguistics.

\bibitem[{Kohli et~al.(2022)Kohli, Kaur, and Bedi}]{kohli2022adversarial}
Guneet~Singh Kohli, Prabsimran Kaur, and Jatin Bedi. 2022.
\newblock {Adversarial Perturbations Augmented Language Models for Euphemism
  Identification}.
\newblock In \emph{Proceedings of the 3rd Workshop on Figurative Language
  Processing}, pages 154--159. Association for Computational Linguistics.

\bibitem[{Lee et~al.(2024)Lee, Chirino~Trujillo, Cuevas~Plancarte, Ojo, Liu,
  Shode, Zhao, Feldman, and Peng}]{lee-etal-2024-meds}
Patrick Lee, Alain Chirino~Trujillo, Diana Cuevas~Plancarte, Olumide Ojo, Xinyi
  Liu, Iyanuoluwa Shode, Yuan Zhao, Anna Feldman, and Jing Peng. 2024.
\newblock \href {https://aclanthology.org/2024.findings-eacl.59/} {{MED}s for
  {PET}s: Multilingual euphemism disambiguation for potentially euphemistic
  terms}.
\newblock In \emph{Findings of the Association for Computational Linguistics:
  EACL 2024}, pages 875--881, St. Julian's, Malta. Association for
  Computational Linguistics.

\bibitem[{Lee and Feldman(2024)}]{lee2024multilingual}
Patrick Lee and Anna Feldman. 2024.
\newblock {Report on the Multilingual Euphemism Detection Task}.
\newblock In \emph{Proceedings of the 4th Workshop on Figurative Language
  Processing}, pages 110--114. Association for Computational Linguistics.

\bibitem[{Lee et~al.(2022{\natexlab{a}})Lee, Feldman, and Peng}]{lee2022report}
Patrick Lee, Anna Feldman, and Jing Peng. 2022{\natexlab{a}}.
\newblock \href {https://doi.org/10.18653/v1/2022.flp-1.27} {A report on the
  euphemisms detection shared task}.
\newblock In \emph{Proceedings of the 3rd Workshop on Figurative Language
  Processing (FLP)}, pages 184--190, Abu Dhabi, United Arab Emirates (Hybrid).
  Association for Computational Linguistics.

\bibitem[{Lee et~al.(2022{\natexlab{b}})Lee, Gavidia, Feldman, and
  Peng}]{lee-etal-2022-searching}
Patrick Lee, Martha Gavidia, Anna Feldman, and Jing Peng. 2022{\natexlab{b}}.
\newblock \href {https://doi.org/10.18653/v1/2022.unimplicit-1.4} {Searching
  for {PET}s: Using distributional and sentiment-based methods to find
  potentially euphemistic terms}.
\newblock In \emph{Proceedings of the Second Workshop on Understanding Implicit
  and Underspecified Language}, pages 22--32, Seattle, USA. Association for
  Computational Linguistics.

\bibitem[{Lee et~al.(2023)Lee, Shode, Trujillo, Zhao, Ojo, Plancarte, Feldman,
  and Peng}]{lee-etal-2023-feed}
Patrick Lee, Iyanuoluwa Shode, Alain Trujillo, Yuan Zhao, Olumide Ojo, Diana
  Plancarte, Anna Feldman, and Jing Peng. 2023.
\newblock \href {https://doi.org/10.18653/v1/2023.starsem-1.38} {{FEED} {PET}s:
  Further experimentation and expansion on the disambiguation of potentially
  euphemistic terms}.
\newblock In \emph{Proceedings of the 12th Joint Conference on Lexical and
  Computational Semantics (*SEM 2023)}, pages 437--448, Toronto, Canada.
  Association for Computational Linguistics.

\bibitem[{Pires et~al.(2019)Pires, Schlinger, and Garrette}]{mbert}
Telmo Pires, Eva Schlinger, and Dan Garrette. 2019.
\newblock \href {https://doi.org/10.18653/v1/P19-1493} {How multilingual is
  multilingual {BERT}?}
\newblock In \emph{Proceedings of the 57th Annual Meeting of the Association
  for Computational Linguistics}, pages 4996--5001, Florence, Italy.
  Association for Computational Linguistics.

\bibitem[{{Real Academia Española}(2025)}]{rae2025corpes}
{Real Academia Española}. 2025.
\newblock \href {https://apps2.rae.es/CORPES/view/inicioExterno.view} {{CORPES
  XXI: Corpus del Español del Siglo XXI}}.
\newblock Accessed: 2025-05-25.

\bibitem[{Vitiugin and Paakki(2024)}]{vitiugin2024ensemble}
Fedor Vitiugin and Henna Paakki. 2024.
\newblock {Ensemble-Based Multilingual Euphemism Detection: A Behavior-Guided
  Approach}.
\newblock In \emph{Proceedings of the 4th Workshop on Figurative Language
  Processing}, pages 73--78. Association for Computational Linguistics.

\bibitem[{Wang et~al.(2022)Wang, Liu, Zhang, Fan, and Guo}]{wang2022euphemism}
Yuting Wang, Yiyi Liu, Ruqing Zhang, Yixing Fan, and Jiafeng Guo. 2022.
\newblock {Euphemism Detection by Transformers and Relational Graph Attention
  Network}.
\newblock In \emph{Proceedings of the 3rd Workshop on Figurative Language
  Processing}, pages 79--83. Association for Computational Linguistics.

\bibitem[{Wiegand et~al.(2023)Wiegand, Kampfmeier, Eder, and
  Ruppenhofer}]{wiegand2023euphemistic}
Michael Wiegand, Jana Kampfmeier, Elisabeth Eder, and Josef Ruppenhofer. 2023.
\newblock {Euphemistic Abuse – A New Dataset and Classification Experiments
  for Implicitly Abusive Language}.
\newblock In \emph{Proceedings of the 2023 Conference on Empirical Methods in
  Natural Language Processing}, pages 16280--16297. Association for
  Computational Linguistics.

\bibitem[{Zhu et~al.(2021)Zhu, Gong, Bansal, Weinberg, Christin, Fanti, and
  Bhat}]{zhu-2021-self-supervised}
Wanzheng Zhu, Hongyu Gong, Rohan Bansal, Zachary Weinberg, Nicolas Christin,
  Giulia Fanti, and Suma Bhat. 2021.
\newblock \href {https://arxiv.org/abs/2103.16808} {Self-supervised euphemism
  detection and identification for content moderation}.
\newblock \emph{CoRR}, abs/2103.16808.

\end{thebibliography}
\end{document}